\documentclass[runningheads]{llncs}
\pdfoutput=1

\usepackage{amsmath}
\usepackage{graphicx} 
\usepackage{tabularx} 
\usepackage{booktabs} 
 
\usepackage{eccv}

\usepackage{eccvabbrv}

\usepackage{graphicx}
\usepackage{booktabs}

\usepackage[accsupp]{axessibility}  

\usepackage[pagebackref,breaklinks,colorlinks,citecolor=eccvblue]{hyperref}

\usepackage{orcidlink}

\begin{document}

\title{Saliency-Aware: Automatic Buddha Statue Recognition}

\author{Yong Qi\inst{1,2}\thanks{Corresponding author: \email{qiyong@sust.edu.cn}} \and
Fanghan Zhao\inst{1,2}}

\institute{School of Electronic Information and Artificial, Shaanxi University of Science and Technology, 710021, Xi'an, China \and
Shaanxi Joint Laboratory of Artificial Intelligence, Shaanxi University of Science and Technology, 710021, Xi'an, China}

\maketitle

\begin{abstract}
	Buddha statues, as a symbol of many religions, have significant cultural implications that are crucial for understanding the culture and history of different regions, and the recognition of Buddha statues is therefore the pivotal link in the field of Buddha study. However, the Buddha statue recognition requires extensive time and effort from knowledgeable professionals, making it a costly task to perform. Convolution neural networks (CNNs) are inherently efficient at processing visual information, but CNNs alone are likely to make inaccurate classification decisions when subjected to the class imbalance problem. Therefore, this paper proposes an end-to-end automatic Buddha statue recognition model based on saliency map sampling. The proposed Grid-Wise Local Self-Attention Module (GLSA) provides extra salient features which can serve to enrich the dataset and allow CNNs to observe in a much more comprehensive way. Eventually, our model is evaluated on a Buddha dataset collected with the aid of Buddha experts and outperforms state-of-the-art networks in terms of Top-1 accuracy by 4.63\% on average, while only marginally increasing MUL-ADD.
\end{abstract}

\section{Introduction}
\label{sec:intro}

The recognition of Buddha statues has always been a difficult and costly task, especially due to the long history of China and the Silk Road in recent years, resulting in a wide variety of Buddha statues. Even Buddha statues of the same kind can vary greatly in construction methods  and material selection due to a dramatic regional and cultural divergence. As such, the accurate identification of Buddha statues can be difficult, even for experienced Buddha experts.

With the development of computer technology in recent years, many new technologies have emerged, such as machine learning and convolutional neural network. These new technologies have solved some problems that were difficult or unsolvable in the past. For example, a simple stacked convolution neural network\cite{lecun1989backpropagation} can easily outperform hand-crafted operators such as the SIFT operator \cite{lowe2004distinctive} by a significant margin. The selected inductive bias equips convolution neural networks with robust abilities for visual information extraction. Therefore, how to effectively combine convolution neural networks with Buddha statue recognition has become a crucial point.
		
Previous works \cite{castellano2022leveraging, renoust2019historical, garcia2019context} explored how to efficiently combine the strength of convolution neural networks(CNNs) with artwork analysis. The results of these attempts demonstrate the impressive power of convolution neural networks in the field of artwork analysis. All of these works, however, involve the construction of knowledge graphs in order to introduce extra information into the network, which has proven to be an effective way to improve classification accuracy, but with significantly more effort and resources involved. We argue that the maximum use of visual information can also greatly improve the classification performance, with the scale of trainable parameters of the network remaining unchanged and only a marginal increase in computation. Therefore, in this paper, we introduce the Grid-wise Local Attention(GLSA) module to extract saliency regions of images in order to enrich the visual information presented to the network. In the end, we experiment our proposed module on a Buddha statue dataset, and the eventual results demonstrate the effectiveness of GLSA.

The rest of the paper is organized as follows:  In section \ref{two} related works are introduced; In section \ref{three} the details and observations of the dataset; Section \ref{four} contains the detailed explanation of the proposed module; In section \ref{five} the experiment details are introduced; In section \ref{six} the final conclusion is presented.

\section{Related Work}\label{two}
{\bfseries{Automatic art analysis}} is a technology rising with the development of data science in recent years. It aims to use machine learning and deep learning to analyze various characteristic data of artworks, so as to determine the author, creation year and value of artworks.
	
Previous researches\cite{carneiro2012artistic, johnson2008image, khan2014painting, mensink2014rijksmuseum} utilize hand-crafted operator to run artwork analysis. All these hand-crafted can perform exceptionally on some specific tasks. But the inherent hand-crafted nature endows them with poor generalization performance. This naturally leads to the employ of deep learning techniques in this field due to its excellent performance in generalization and simplicity. 
	
Giovanna \emph{et al}.\cite{castellano2022leveraging} takes the advantage of the graph-based method in order to inject context information into the model, which indeed improves the classification performance but with a high cost of the construction of context feature embedding graph. We argue that the performance of classification models can be improved with complete and comprehensive exploits of visual features, which leads us to saliency map extraction.
	
{\bfseries{Saliency-guided CNNs}} Visualizing the salient regions of an image\cite{adebayo2018sanity,ribeiro2016should,selvaraju2017grad,simonyan2013deep,springenberg2014striving,zeiler2014visualizing} was originally designed to provide an insight into the CNNs'black-box mechanism. Simonyan \emph{et al}. \cite{simonyan2013deep} proposed an approach that computes importance scores for every pixel through back propagation based on Taylor expansion and the resultant saliency maps exhibit which areas of the target image are of the most significance in the classification process. Furthermore, the Deconvolution-based method\cite{zeiler2014visualizing} was proposed with reverse operations of filtering, pooling and activation layers.  Class Activation Mapping(CAM)\cite{zhou2016learning} and Grad-CAM\cite{selvaraju2017grad} are superior to Gradient-based methods\cite{simonyan2013deep} in being highly class-discriminative whereas the latter are better at retaining high-resolution details, therefore, Guided Grad-CAM combines the strength of both world and provides a high-resolution class-discriminative visualization.
\begin{figure*}[tpb] 
    \centering  
    \includegraphics[width=1\linewidth, height=0.4\linewidth]{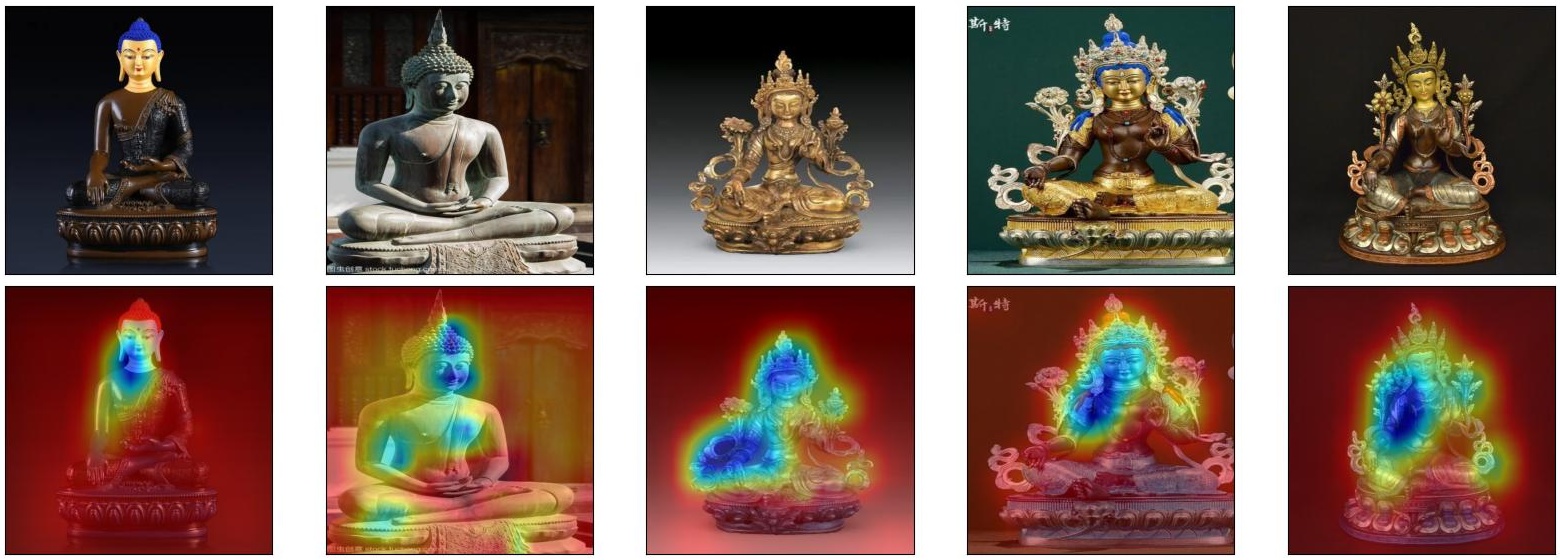}
    \caption{Saliency maps with ResNet-18 as the basic convolution unit. We can observe a pattern where salient features mostly surround the key part of the Buddha statue's body, such as the face and hand.}
    \label{Saliency}
\end{figure*}	

	Inspired by recent saliency-guided data augmentation\cite{kim2020puzzle,uddin2020saliencymix,choi2021salfmix,huang2021snapmix}, which enhance the expressivity of the dataset by cropping the saliency region of the source image and pasting it to the target image. We devise a saliency-guided sampling model, which, in other words, samples a fixed number of cropped images with a fixed size via a pre-trained model and then fed forward into standard CNNs to obtain visual feature embedding. Eventually, the visual feature embedding of the saliency branch is concatenated with the visual feature embedding from another branch whose task is to comprehend the image globally. The final feature embedding is fed into a linear projection layer to make the classification decision.

	{\bfseries{CNN and Vision transformer}} Convolution neural networks (CNNs)\cite{krizhevsky2012imagenet, lecun1989backpropagation} have proven to be a highly effective approach in handling visual information, It is secure to say its correct selections of inductive bias, namely translation invariance and spatial bias, earn the most credits. However, its inductive biases also limit its potential and ability to address global visual features.
	
	Vision Transformer(ViT)\cite{dosovitskiy2020image} is proposed with the introduction of attention mechanism\cite{vaswani2017attention} into computer vision, which endows vision models with global feature processing ability.
	In this work, we proposed a Grid-wise Local Self-Attention(GLSA) module which combines the strength of ViT in global feature processing with the depth-wise property of MobileNet\cite{howard2017mobilenets, sandler2018mobilenetv2} to refine and magnify regional salient feature response, which eventually proven to be an effective method in improving the classification performance.

	Our contributions are:
	\begin{itemize}
		\item To propose a Grid-wise Local Self-Attention(GLSA) module that enhances and magnifies the salient feature response
		\item To provide a way of salient feature map sampling that improves the classification performance with only a little increase in MUL-ADD
	\end{itemize}

\section{Statistics and Observation}\label{three}
{\bfseries{Dataset details}} the collection of Buddha images was carried out under the guidance of Buddha experts.  The final Buddha dataset consists of two parts. The first part, which we denote as $\Re_1$,  is a wide range of Buddha statues from different religion, region and culture backgrounds, and is used dedicatedly in the training process of the Grid-wise Local Self-Attention(GLSA)  module, and later is frozen to serve merely as a salient feature extraction function.

\begin{figure}[h]
    \centering
    \includegraphics[width=1\linewidth, height=0.8\linewidth]{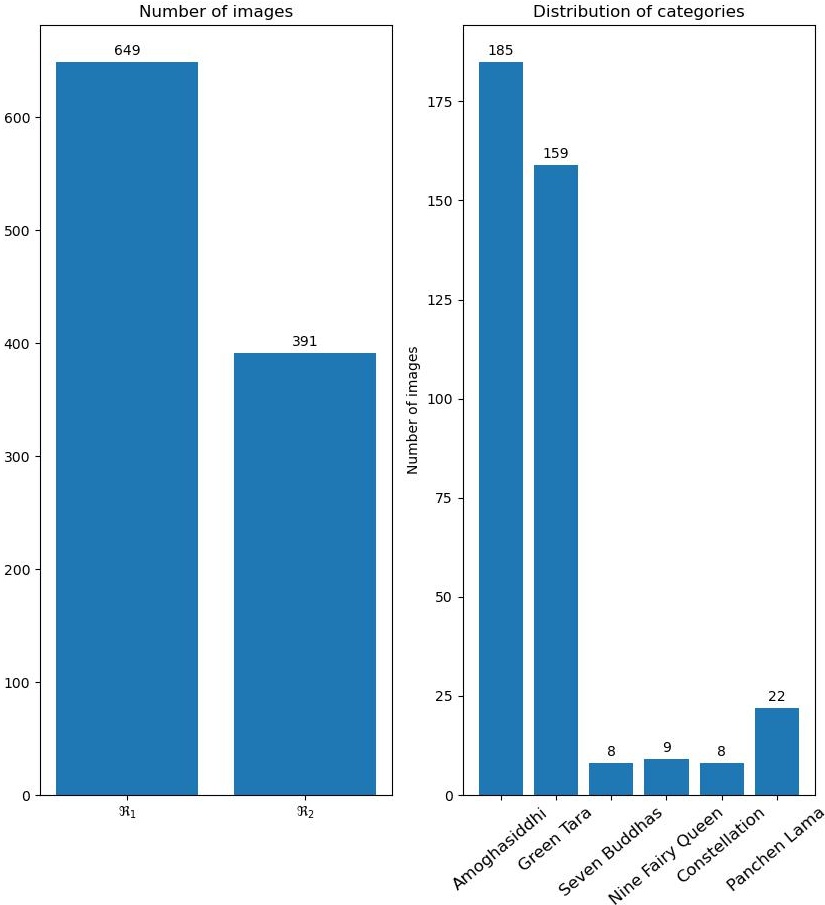}
    \vspace{-2em}
    \caption{Details of dataset}
    \label{dod}
\end{figure}

The second part, denoted as $\Re_2$, consists of six categories of Buddha statues, namely Amoghasiddhi,  Green Tara, Seven Buddha,  Nine Fairy Queen,  Constellation and Panchen Lama. Fig. \ref{dod} illustrates the details of the dataset.

{\bfseries{Observation and Analysis}} we firstly trained a standard Resnet18\cite{he2016deep} on $\Re_1$ and attempted to analyze which areas of images are highly activated in the process of classification by using the gradient-based salient feature visualization method proposed in \cite{simonyan2013deep}.  which can be briefly summarized as follows:
\begin{align}
    S_c &= f(I;\theta) \\  
    Grad(I)&= \left. \frac{\partial S_c}{\partial I} \right\rvert_I \label{Grad} \\
    \intertext{Where \(I\) is the input image, \(f(\cdot)\) is a convolution neural network parameterized with \(\theta\). \(S_c\) is the output of the network which in classification tasks is the classification score. \(Grad(\cdot)\) is the importance of each image pixel, which can be formulated:}
    S_c(I) &\approx w^T I + b
\end{align}
Where \(w\) is the derivative of classification score, as explained in \eqref{Grad}

The salient feature maps are presented in Fig.\ref{Saliency} with Resnet18\cite{he2016deep} as the basic convolution network.


The pattern we can observe from Fig.\ref{Saliency} is that the highly activated areas are mainly concentrated around the face and hands of Buddha statues, meaning that these regions are the salient regions of the image and a slight change in these regions can translate into a relatively larger change in the classification score than in other regions. Also, these salient visual features can serve as an effective source in expanding the dataset and relieving the class imbalance problem, and further improving the classification performance.
 
\begin{figure}[!h] 
    \centering
    \includegraphics[width=0.75\linewidth, height=0.6\linewidth]{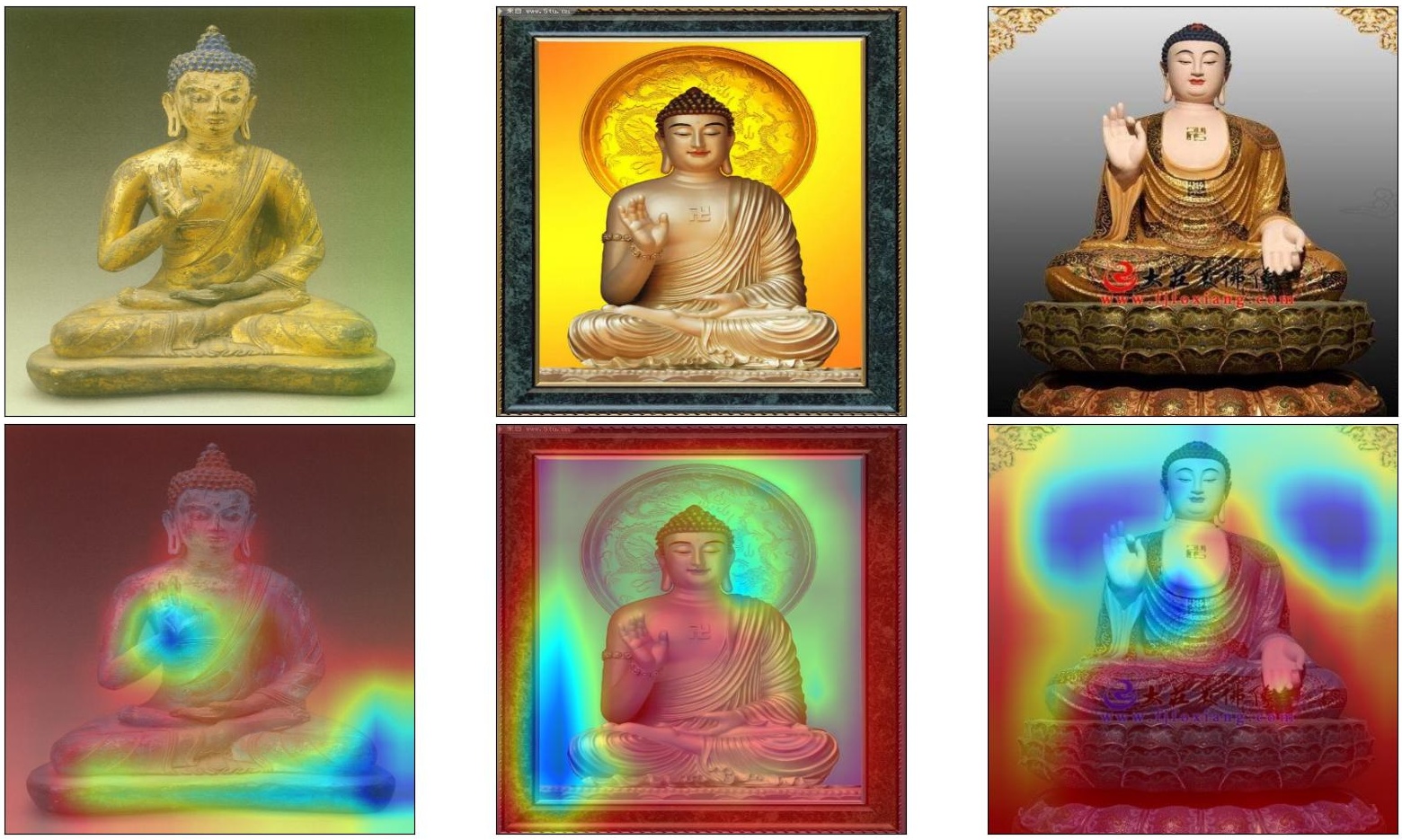}
    \vspace{-1em}
    \caption{Defective saliency maps}
    \label{Bad}
\end{figure}

However, Fig.\ref{Bad} shows us that Resnet is defective in saliency map extraction due to the class imbalance problem and great divergence between the distribution of different images. Therefore, we propose a Grid-wise Local Self-Attention(GLSA) module instead of Resnet as the basic saliency maps extraction module. GLSA is trained on $\Re_1$ which contains the location of salient features obtained from the analysis of saliency map extracted by Resnet. And the task of GLSA is to predict the objectiveness of salient features in each sliced image grid.

\section{Proposed Method} \label{four}
	\subsection{{\bfseries{Overview}}}
\begin{figure*}[h]
	\centering
	\includegraphics[width=0.8\linewidth, height=0.5\linewidth]{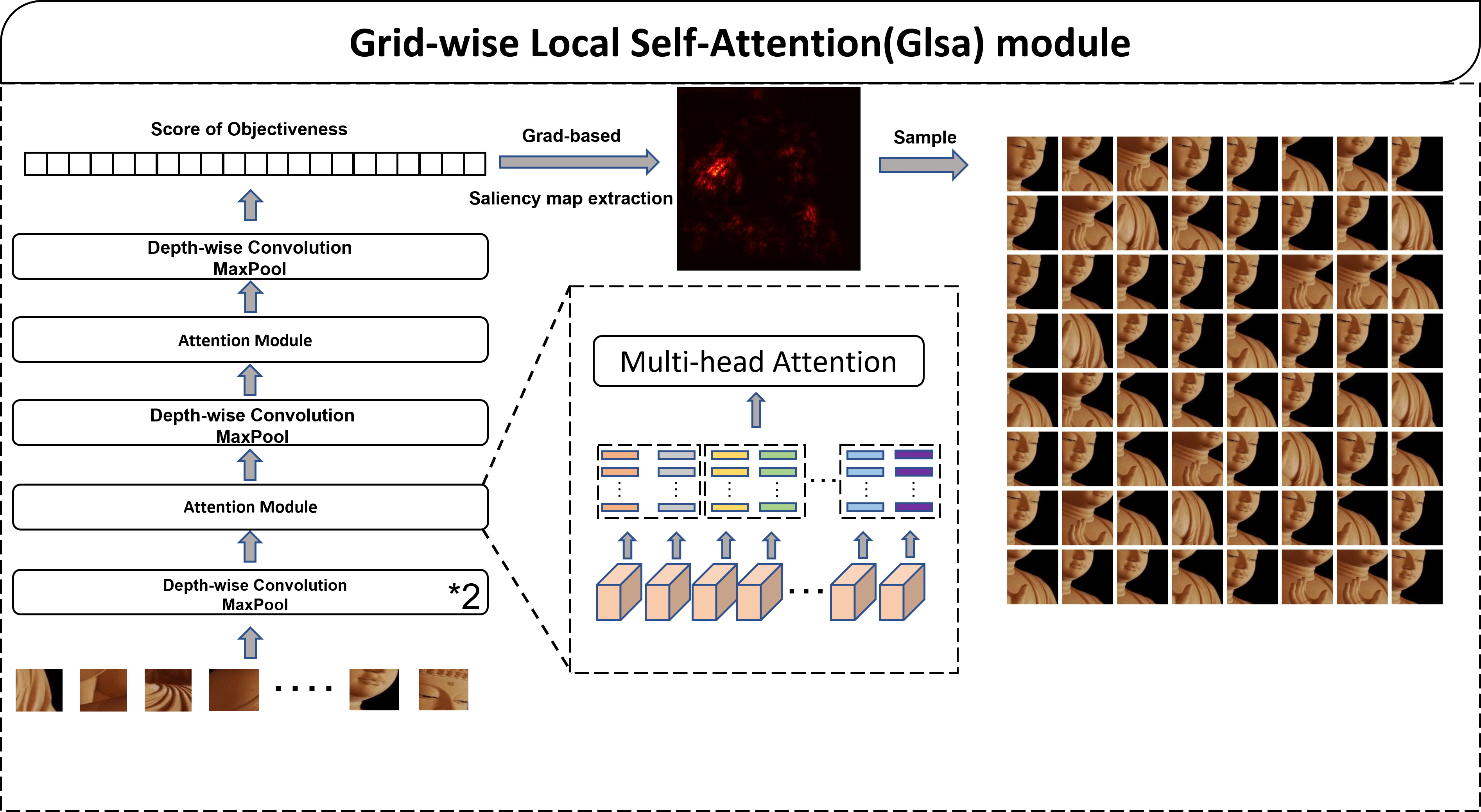}
	\caption{Overview of Grid-wise Local Self-Attention(GLSA) Module}
	\label{GLSA}
	\end{figure*}

	The main challenge in recognizing Buddha statues is that features of different Buddha statues overlap a lot due to regional and cultural differences, and some features of the same Buddha statues are irrelevant. However, we can observe in Fig.\ref{Saliency} that some key regions are highlighted in the classification process. Therefore, given the above properties, our proposed model as shown in Fig.\ref{GLSA} focuses on extracting salient features of an image, but instead using the extracted features in the way like \cite{uddin2020saliencymix, huang2021snapmix, kim2020puzzle}, in which salient features are used to replace part of the original image akin to Cutout\cite{devries2017improved} and Mixup\cite{zhang2017mixup}.  We treat these salient feature maps as activation maps and feed them into a standard convolution network to obtain visual embeddings of extracted salient feature maps. 
	
	\begin{figure*}[tb]
	\centering
	\includegraphics[width=0.8\linewidth, height=0.5\linewidth]{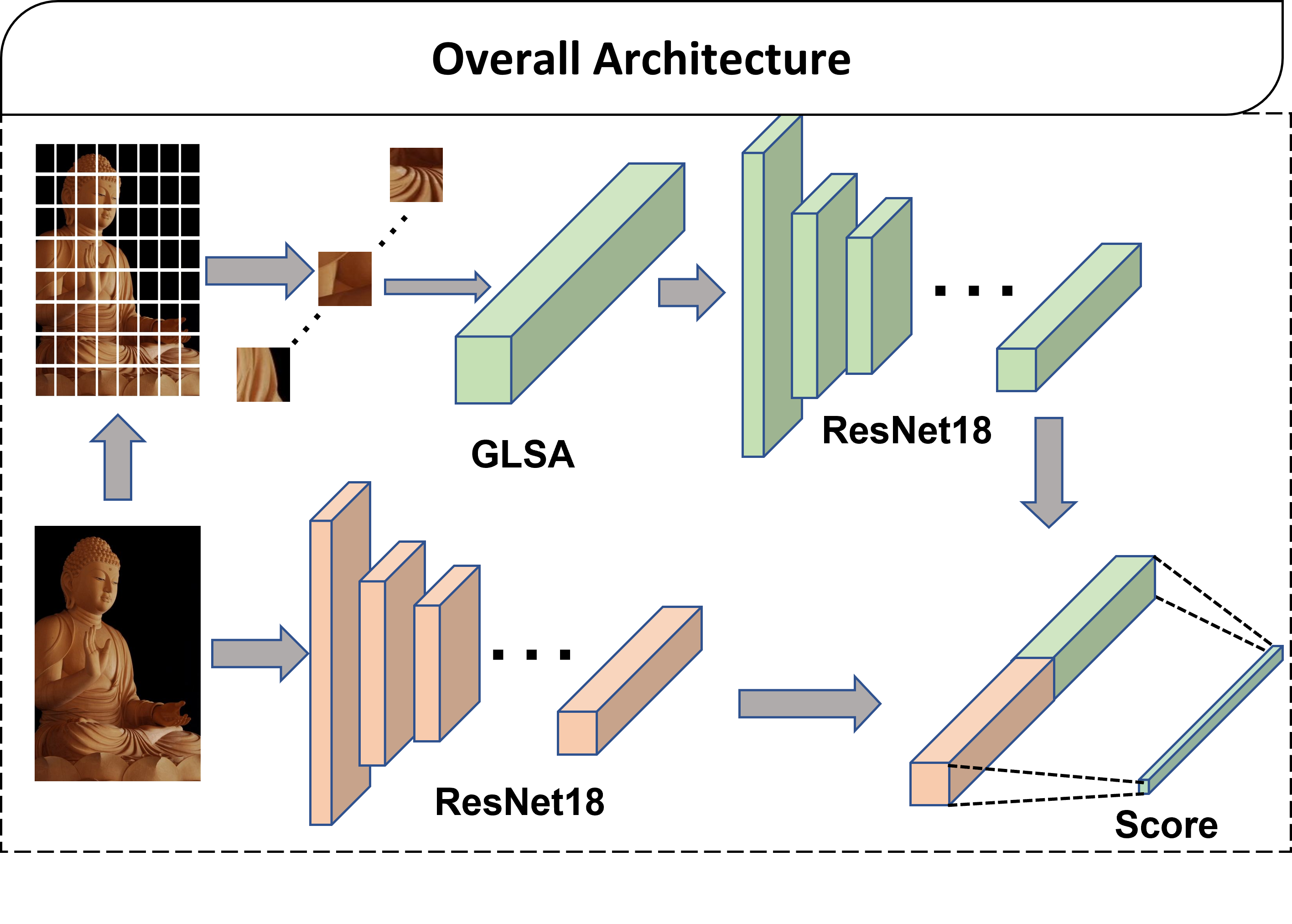}
	\vspace{-2em}
	\caption{Overview of the classification model}
	\label{Stru}
\end{figure*}	  
	To be more specific, for a given image of size \(3 \times D_w \times D_h\) where \(D_w\) and  \(D_h\)  are the width and height of an image,  we first feed it into the GLSA module, which consists of depth-separable convolution modules and self-attention modules, to obtain a sliced image with a size of \(S_n \times S_w \times S_h \) , where  \(S_n\) is a predetermined parameter that decides the number of salient feature maps,  \(S_w\) and \(S_h\) are the pre-determined width and height of extracted feature map. The retrieved salient feature maps are then fed into a standard convolution network to obtain visual embedding. and eventually are combined with the visual embedding from another path whose responsibility is to process an entire image and provide a global and comprehensive understanding of the image to make the classification together.
	
	\subsection{{\bfseries{Grid-wise Local Self-Attention Module}}}
	
	To better comprehend the attention mechanism used in GLSA module, we will first address the partition strategy we employed. As we have observed in Fig.\ref{Bad}, Simply Resnet or any other classification model might potentially generate some outlier pixels which poses a potential hazard in retrieving salient feature maps and ultimately limit the distribution of salient feature maps

	Therefore, GLSA firstly slices one image into  \(3N^2 \times \frac{D_w}{N} \times \frac{D_h}{N}\) where \(N\) is a pre-determined parameter for the number of grids into which we divide the image and then these grids are flattened and eventually fed into following convolution modules.
	
	{\bf{\emph{Depth-separable convolution}}} is selected for its excellence in processing feature maps of different channels separately and concurrently. To make this easier to understand, our goal for depth-separable convolution is to detect the objectiveness of salient features in each grid, instead of performing classification which requires the entire image to be considered at once.\\ 
	\indent
	{\bf{\emph{Attention module}}} is introduced into GLSA to promote local information exchange.  We observed in experiments that the zero-information-exchange strategy does not deliver very satisfactory results, instead the enabling of local information exchange can greatly improve the salient feature detection performance. Therefore, we introduced a new variable Proximity, denoted as \(\S \), into the attention module(Fig.\ref{GLSA}) to control the extent of the vicinity.
	Proximity is defined as the number of depth-separable entities contained in one group. Depth-separable entity refers to a group of feature maps derived from the same sliced image patch.  Detailed structure is depicted in Fig.\ref{GLSA}.  The attention module can be formulated as:
	\setlength\abovedisplayskip{0.3cm}
	\setlength\belowdisplayskip{0.3cm}
	\begin{gather}
		Out = concat(Attention(F_1),\ldots,Attention(F_g)) \\
		Attention(F_i)=concat(head_1,head_2,\ldots,head_{N_h})
	\end{gather}
	Where \(F_i\)  is the feature map and \(g=\frac{Channels}{\S}\). As stated above, proximity refers to how many channels we take into consideration in one attention module, and therefore \(g\) stands for the number of groups. \(N_h\)  is the number of different heads. \(concat\) denotes the concatenation operation.
	We use multi-head self-attention(MHSA) to process the information in one group. Details of MHSA can be formulated as follows:
\begin{align}
   head_j &= Attention(Q_j,K_j,V_j)\\
          &= Softmax\left(\frac{Q_jK_j^T}{\sqrt{d}}\right)V_j\\
          &= Softmax\left(\frac{F_jW_j^QF_jW_j^K}{\sqrt{d}}\right)F_jW_j^V
\end{align}

Where \(Q_j=F_jW_j^Q\) ,\(K_j=F_jW_j^K\), \(V_j=F_jW_j^V\).
	 
	{\bfseries{Training}}. In the training of the GLSA module, we aim to maximize the response of salient features in different regions. Therefore, for a given input \(X\) , the target is built based on:
	\begin{align} \label{Saliency extraction}
		X^o &= ConNet(X)  \\
		X^{BP} &= BP(X^o) \\
		Target &= Thresh(\lbrace X_j^{BP} \rbrace _{j=1}^{N_g})  
	\end{align}
	Where \(ConNet(\cdot)\) is any convolution neural network and ResNet is selected in our case.\(ConNet(\cdot)\) helps to get the objectiveness prediction of each grid, and then an average operation is performed over positive predictions to obtain \(X^o\). The \(BP(\cdot)\) operation is based on \cite{simonyan2013deep}, and aims to extract the salient feature maps. The \(Thresh(\cdot)\) operation sets the objectiveness of grids to 1 if the objectiveness is greater than the chosen threshold. And \(N_g\) stands for the number of grids
	
	The constructed \(\lbrace X_j, Target_j \rbrace _{j=1}^N\) then are fed into the GLSA module and the following optimization is performed:
	\begin{equation}
		\underset{W}{argmin} \sum_{i=1}^N \ell(f(X_i;W), Target_i)
	\end{equation}
	Where \(f\) is the GLSA module with parameters \(W\) ,  \(\ell\) is a Binary Cross Entropy loss function.

	\subsection{{\bfseries Classification Model}} 
	The classification model we have selected is intended for two different levels of computation. One is to process global visual information, in other words, the input is the entire image. While the another is to process sliced inputs from the GLSA module. And the final outputs of both paths are concatenated and fed into a linear projection function to perform the classification. Structure details are shown in Fig.\ref{Stru}:
	
	Formally, the whole classification process can be formulated:
	\begin{align}
		X^{local} &=ConvNet^1(GLSA(X)  \\
		X^{global} &=ConvNet^2(X)\\
		Score &=Softmax(\underset{W_{linear}}{Linear}([X^{local}, X^{global}])) 
	\end{align}
	Where, \(ConvNet^1(\cdot)\) and \(ConvNet^2(\cdot)\) can be any convolution network. \(Linear(\cdot)\) operation is parameterized with vector \(W_{linear}\)  and maps the concatenated results to the number of classes.

	\section{Evaluation} \label{five}

	To evaluate the effectiveness of the proposed network, the experiment is conducted on $\Re_2$ , which is collected for the dedicated use of Buddha statue recognition. Please see Section.\ref{three} for more details.

\begin{table}[ht]
\centering
\caption{Comparison between different models on dataset $\mathcal{R}$. DBN: Dual Branch Network, NTPs: Non-Trainable Parameters, TPs: Trainable Parameters.}
\resizebox{\textwidth}{!}{
\begin{tabular}{|l|*{6}{c|}c|c|c|c|}
\hline
\textbf{Model} & \textbf{C1} & \textbf{C2} & \textbf{C3} & \textbf{C4} & \textbf{C5} & \textbf{C6} & \textbf{Top-1 Acc.(\%)} & \textbf{MUL-ADD(G)} & \textbf{NTPs(MB)} & \textbf{TPs(MB)} \\
\hline
ResNet-18 & 95.23 & 94.67 & 72.41 & 93.45 & 97.64 & 78.32 & 94.43 & 18.95 & 0 & 11.2\\
ResNet-50 & 92.86 & 93.65 & 69.81 & 91.28 & 96.59 & 66.27 & 91.57 & 42.71 & 0 & 23.5 \\
ResNext50-32 & 91.30 & 92.67 & 70.23 & 90.98 & 96.43 & 67.35 & 91.43 & 44.18 & 0 & 23.0\\
EfficientNet-b0 & 92.86 & 94.65 & 71.53 & 92.39 & 96.55 & 76.31 & 92.41 & 4.0 & 0 & 4.0\\
\hline
DBN & 96.30 & \textbf{100.00} & \textbf{88.86} & \textbf{100.00} & \textbf{100.00} & 95.46 & \textbf{97.16} & 37.90 & 0 & 22.9 \\
DBN-GLSA & \textbf{96.35} & \textbf{100.00} & 86.23 & \textbf{100.00} & \textbf{100.00} & \textbf{96.13} & 97.09 & \textbf{20.58} & 4.6 & 22.9\\
\hline
\end{tabular}
}
\label{tab:comparison}
\end{table}

\begin{figure*}[htbp]
	\centering
	\includegraphics[width=0.8\textwidth]{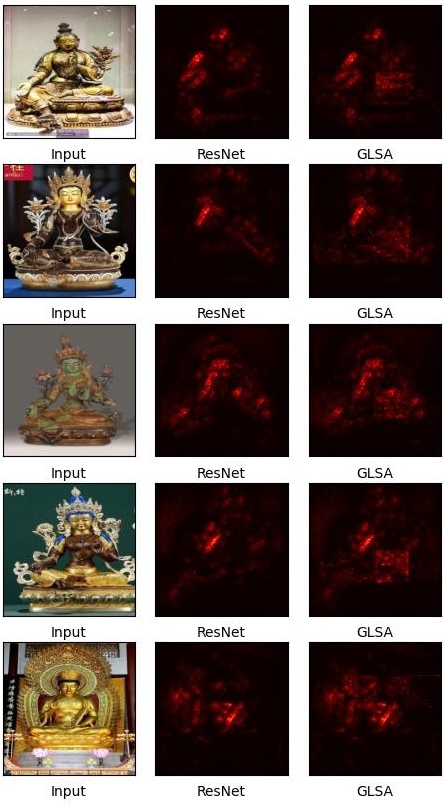} 
	\caption{The comparison between ResNet and GLSA as the saliency map extraction module respectively}
	\label{Comp}
\end{figure*}

	\subsection{Salient Feature Extraction Module(SFEM)}
	The salient feature extraction modules perform calculations stated in \eqref{Saliency extraction}. The comparison experiment is conducted between two different structured modules. Results are shown in Fig.\ref{Comp}:
	
	{\bfseries{GLSA}} is based on channel-separable convolution and attention module and produces the objectiveness of salient features in each grid. And then the sampling of salient feature maps is performed based on the objectiveness score. 
	
	{\bfseries{Resnet}} is proposed in \cite{he2016deep} and has proven to be an effective way of processing visual information. But when applied in extracting salient visual features, it tends to detect features that are not general over the dataset, which eventually translates into biased sampling outcomes.

	\subsection{Baselines}
	The structure of our proposed network is a dual-branch convolution network, as shown in Fig.\ref{Stru}. One of the two branches is responsible for processing salient feature maps, while the another is responsible for processing global visual information. And the resultant visual embeddings of two branches are concatenated to perform the final classification. Therefore, the setting of baseline are:
	
	{\bfseries{Single branch convolution network}} We compare against ResNet\cite{he2016deep}, ResNext\cite{xie2017aggregated}, and EfficientNet\cite{tan2019efficientnet}, which are all currently the state of the art network. We do not load these networks as pre-trained. The initialization scheme employed here is kaiming initialization\cite{he2015delving}. There is only one branch throughout the classification process, which means no extra information is added to the final classification decision.
	
	{\bfseries{Dual-branch convolution network}}  We construct two dual-branch convolution networks in this experiment. One is the proposed network. The another is with almost the same structure, but the GLSA module is removed from the network. This leaves two parallel ResNet along to perform the classification. 
	The two dual-branch networks are also initialized with kaiming initialization.
	\subsection{Result Analysis}

	We compare the performance of each model from three aspects: TOP-1 accuracy of each category, the overall TOP-1 accuracy and the amount of MUL-ADD.
	
	From Table \ref{tab:comparison}, We can observe that the dual-branch network has a significant performance improvement in dealing with the class imbalance problem, especially the accuracy for Green Tara, Nine Fairy Queen and Constellation reached 100\%.  This shows that dual-branch networks are better at extracting visual features than networks with only one branch. Another observation is that ResNet50 with a similar scale of parameters to DBN, but is significantly inferior in classification performance. Our guess is that the model capacity of ResNet50 is so large that leads to overfitting. While the overfitting problem is resolved for dual-branch structured networks, due to the collective processing of  the outputs from two different branches, which acts as a regularization constraint.

	The second observation is that the performance of the network without GLSA is slightly better than using GLSA, but with the cost of twice as much MUL-ADD. A potential explanation is that the salient feature maps extracted using GLSA modules are subsets of global features, which explains the superior performance of the network without GLSA module. However, considering the fact that only serval particular regions of an image are critical and GLSA functions as a salient feature sampler that retrieves salient feature maps from highly activated regions, this leads to a significant reduction in the amounts of MUL-ADD and eventually achieved an increase of 2.66\% in TOP-1 accuracy with only negligible increase in computation requirements, when compared to ResNet18.

\section{Conclusion} \label{six}
	This work presents a Grid-wise Local Self-Attention module for enhancing and magnifying salient feature responses. The comparison between different baselines is performed and the results demonstrate the use of dual-branch networks along with GLSA module can significantly increase the classification performance with no large spikes in computational requirements. In future work, we will further combine different sampling strategies with saliency maps to refine the proposed method.

\bibliographystyle{splncs04}
\bibliography{reference}

\end{document}